\documentclass{article}

\PassOptionsToPackage{numbers, compress}{natbib}
\usepackage[final]{mlsafety_neurips_2022}

\usepackage[utf8]{inputenc} 
\usepackage[T1]{fontenc}    
\usepackage{hyperref}       
\usepackage{url}            
\usepackage{booktabs}       
\usepackage{amsfonts}       
\usepackage{nicefrac}       
\usepackage{microtype}      
\usepackage{xcolor}         
\usepackage{graphicx}
\usepackage{comment}
\usepackage{amsmath,amssymb} 
\usepackage{times}
\usepackage{epsfig}
\usepackage{amsthm}

\usepackage{caption}
\usepackage{subcaption}
\usepackage{wrapfig}
\title{Context-Adaptive Deep Neural Networks via Bridge-Mode Connectivity}

\author{
Nathan Drenkow$^1$~~~~~Alvin Tan$^{1,2}$~~~~~Chace Ashcraft$^1$~~~~~Kiran Karra$^1$\\
$^1$The Johns Hopkins University Applied Physics Laboratory\\
$^2$University of California at Berkeley
}

\begin{document}

\maketitle

\begin{abstract}
  The deployment of machine learning models in safety-critical applications comes with the expectation that such models will perform well over a range of contexts (e.\,g., a vision model for classifying street signs should work in rural, city, and highway settings under varying lighting/weather conditions). However, these one-size-fits-all models are typically optimized for average case performance, encouraging them to achieve high performance in nominal conditions but exposing them to unexpected behavior in challenging or rare contexts.  To address this concern, we develop a new method for training context-dependent models.  We extend Bridge-Mode Connectivity (BMC)~\cite{garipov2018loss} to train an infinite ensemble of models over a continuous measure of context such that we can sample model parameters specifically tuned to the corresponding evaluation context.  We explore the definition of \emph{context} in image classification tasks through multiple lenses including changes in the risk profile, long-tail image statistics/appearance, and context-dependent distribution shift. We develop novel extensions of the BMC optimization for each of these cases and our experiments demonstrate that model performance can be successfully tuned to context in each scenario.  
\end{abstract}

\section{Introduction}
Machine learning (ML) systems must be capable of operation in a wide range of contexts and scenarios to be effective in safety-critical deployments.  While system-level safeguards can constrain an ML model's behavior to ensure safe operation, it is desirable that the model itself would implicitly (through its design or optimization) learn to make safe predictions.  Current conventions for training deep networks assume that trained models will perform well over all operating conditions. However, this one-size-fits-all approach introduces limitations that have been demonstrated across a variety of tasks and benchmark datasets~\cite{hendrycks2019benchmarking, taori2020measuring, kamann2021benchmarking, michaelis2019benchmarking}, raising concerns about the ability of any single model to perform well over many contexts.  In this work, we consider an orthogonal perspective: is it possible to maintain or improve model performance without retraining via adaption according to the model's specific operating context at test-time?

As a motivating example, consider an autonomous vehicle which must be able to detect and classify street signs. The vehicle's \emph{context} may be approximated by measurements such as its speed (e.\,g. accelerating vs. braking), location (e.\,g. rural vs. urban), and/or current environmental conditions (e.\,g. rain vs. sunshine).  In these cases, the context is measurable and exists along a continuum. More importantly, the context provides additional information about, e.\,g., class priors or critical failure modes.  For instance, if the vehicle is operating at a high speed, then mistaking a \verb|50mph| sign for a \verb|Stop Sign| has more severe consequences than mistaking for a \verb|40mph| sign.  Furthermore, it may also be more likely to observe the \verb|40mph| sign over the \verb|Stop Sign| in that context as well. This illustrative example points to the fact that the system's operating state and context provide useful cues regarding what constitutes safe behavior and critical failure modes.

\textbf{Contributions~~~}
In a paradigm shift away from the conventional one-size-fits-all approach, we develop a method to train models that can be adapted on-the-fly at test-time to different contexts without retraining.  Our approach extends BMC to incorporate three definitions of \emph{context}. We then use the parameterization of the BMC curve to sample a context-specific model for evaluation.

\textbf{Related Work~~~}
\textit{Test-time adaptation}~\cite{sun2019test, wang2020tent, schneider2020improving, zhang2021memo, zhang2021adaptive} considers how model weights, activations, and/or architecture might be modified at test-time to optimize model performance on current test data. While adaptive, these models are informed about context only through data from the testing domain rather than any other measures of system state and potentially without reference to the original training distribution.

\textit{Cost-sensitive robustness}~\cite{khan2017cost, zhang2018cost, zhou2005training} methods consider how to adjust model performance to deal with, e.\,g., class imbalances or to avoid specific failure cases.  While not adaptive at test time, these methods manipulate the training objective to build models which may be implicitly robust to rare classes/scenarios or safety-critical decisions.

\textit{Bridge-Mode Connectivity} BMC~\cite{garipov2018loss} was initially identified as a means to learn paths of equivalent loss in weight-space connecting two pre-trained models. The points along the path correspond to unique sets of interpolated weights such that the learned path can be repeatedly sampled to quickly generate ensembles of neural networks. Beyond ensembling, BMC has been applied in the domain of adversarial machine learning~\cite{zhao2020bridging} for mitigating various forms of train-time attacks. In these cases, the weights for the endpoints are assumed compromised/vulnerable and the bridge-mode enables mitigation via sampling a set of weights from the learned ensemble away from the endpoints that maintains performance but mitigates the vulnerability. To our knowledge, our approach is the first to consider using the bridge-mode parameterization as a proxy for \emph{context} (See Sec.~\ref{sec:context}) and modifying the objective and/or optimization to align the model's performance with the associated context.

\section{Interpretations of Operating Context}
\label{sec:context}
The primary goal of this work is to develop context-dependent models.  We assume context can be measured and represented by a parameter $c \in \mathbb{R}$ and normalized such that $c \in [0, 1]$.  Also, denote $(X, Y) \sim \mathcal{D}$ where $(X, Y)$ are (data, label) pairs drawn from a distribution $\mathcal{D}$. We consider three possible interpretations of \emph{context}.

\textbf{(1) Risk profile -} In real world deployment, context may necessitate that the DNN prioritize certain types of predictions and avoid others. In this sense, we model \emph{context} here by making the loss function dependent on the context parameter: $l(\theta) \longrightarrow l(\theta, c)$. 

\textbf{(2) Long-tail Robustness -} In many deployment scenarios, data may be corrupted naturally by factors due to the environment, sensor, and/or other sources.  
For example, images collected for autonomous driving may be biased toward an expected set of operating conditions, and rare weather events, motion artifacts, noise or other naturally-induced corruptions may be underrepresented in training.
We will assume that $\mathcal{D}$ is stationary but that corrupted samples come from the tails of the distribution. The frequency of corrupted samples occurring as a function of the corruption strength can be captured by the following assumption: if $c' > c$, then $p(c') \leq p(c)$. This limits the types of corruptions to those that may occur naturally (given assumptions about the underlying image generating process) and treats corresponding samples as in-distribution. In this case, we model \emph{context} by a corruption function $\psi(x, c)=x'$.  We assume that $\psi$ obeys the property: $c' > c \Rightarrow |\psi(x,c') - x| > |\psi(x, c) - x|$.   

\textbf{(3) Contextual distribution shift -} Lastly, we consider the case where the data distribution is also parameterized by $c$ such that $(X,Y) \sim \mathcal{D}_c$.  In this case, we define \emph{context} such that $\mathcal{D}_{c'} \neq \mathcal{D}_{c}$ for $c' \neq c$. Since context is assumed to exist along a continuum, this perspective reduces to the context-agnostic view when $\mathcal{D}_{c}$ does not change with $c$ and allows for the difference between $\mathcal{D}_{c'}, \mathcal{D}_{c}$ to be small when $|c' - c|$ is also small (i.e., the distribution can also vary smoothly as $c$ changes). This interpretation captures cases where, e.g., label priors may shift as a function of $c$ (i.e., $p(Y|c) \neq p(Y|c')$ for $c \neq c'$). For example, in the street-sign classification case discussed earlier, city driving increases the likelihood of \verb|Stop Sign| or \verb|Yield| whereas highway driving decreases those same class likelihoods in favor of \verb|70 mph|. 

\section{Bridge-Mode Connectivity}
The original BMC formulation from~\cite{garipov2018loss} learns a low-loss path connecting two sets of pre-trained model weights, $\theta_0$ and $\theta_1$. The method learns a curve $\phi$ parameterized by $t \in [0, 1]$ and an accompanying set of bridge-mode weights $\theta_b$ which interpolate between the endpoints $\theta_0$, $\theta_1$. 

\begin{equation}
    \label{eq:bmc_parametrization}
    \phi_{\theta_b}(t) = (1-t)^2 \theta_0 + 2t(t-1)\theta_b + t^2 \theta_1 
\end{equation}

\noindent The BMC training objective minimizes:

\begin{equation}
    \label{eq:bmc_loss}
    L(\theta_b) = \mathbb{E}_{t\sim\mathcal{U}(0,1)}[l(\phi_{\theta_b}(t)) ]
\end{equation}

\noindent where $l(w)$ is the cross-entropy loss and $w$ is the set of weights of the the BMC network. Standard training proceeds by first sampling a value of $t \sim U(0,1)$, then drawing a minibatch of data, $B_i$, to optimize the BMC loss (\ref{eq:bmc_loss}) for the sampled value of $t$. Here, $i$ indicates the training step index. Note that $B_i$ does not depend on $t$ and in conventional training of deep networks and BMC curves, and typically consists of samples drawn independently from $\mathcal{D}$. The endpoints $\theta_0$ and $\theta_1$ can be fixed or allowed to float while training. 

When adapting the BMC to model the various interpretations of context, we assume $t \equiv c$ in our experiments for simplicity.  However, more general relationships between $t$ and $c$ are not precluded.

\section{Experiments}
\begin{figure}[t!]
\centering
\begin{subfigure}[b]{0.5\textwidth}
  \centering
  \includegraphics[width=0.95\linewidth]{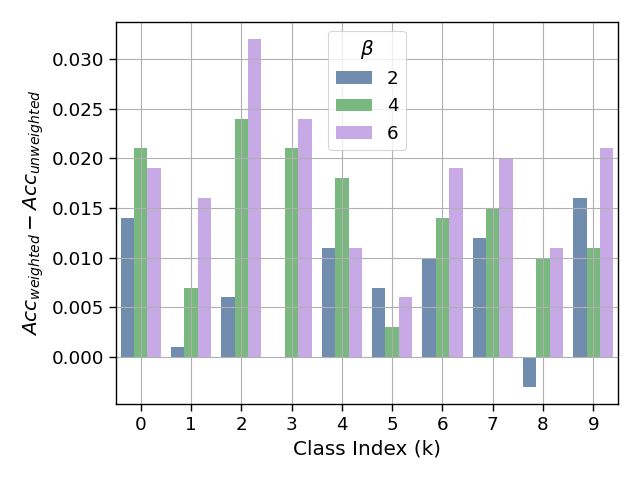}
  \label{fig:exp1}
\end{subfigure}%
\begin{subfigure}[b]{0.5\textwidth}
  \centering
  \includegraphics[width=0.95\linewidth]{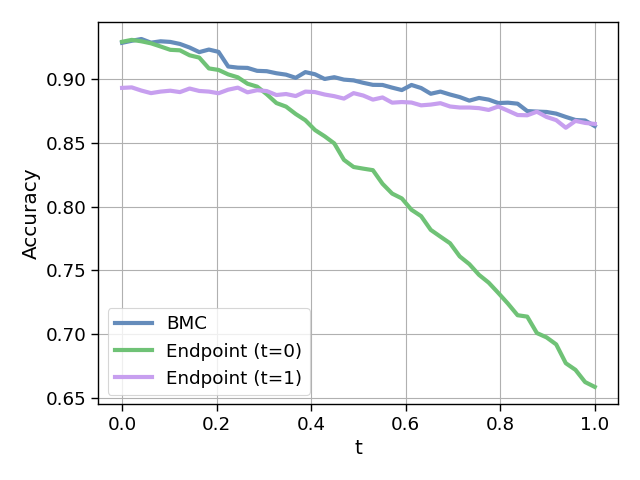}
  \label{fig:exp2}
\end{subfigure}%
\caption{\small\textbf{(left) Adaptive Risk Profile: } Comparing the per-class accuracies of the up-weighted and unweighted BMC models shows that the risk profile is positively shaped such that higher class weights ($\beta$) also produce higher accuracy gains. \textbf{(right) Long-tail Robustness: } The BMC performance is maintained as the corruption severity increases and interpolates between endpoints specialized for clean and heavily-corrupted data.}
\label{fig:exp1_exp2}
\end{figure}

\subsection{Adaptive Risk Profile via Class-Weighted BMC}
Here we use $t \in [0, 1]$ as a proxy for context and first partition $[0, 1]$ into bins, each corresponding to a specific class.  To enable tuning model performance towards particular classes, we change $l$ in (Eq.~\ref{eq:bmc_loss}) to be dependent on the value of $t$ as follows: $l(w, t) = \sum_{k \in K} -\alpha(t, k) \cdot p_k \log p_k$, where $K$ is the set of class labels and $k$ is the index of a particular class. We use $\alpha(t, k)$ to up-/down-weight specific classes in the cross-entropy (CE) loss in order to tune the BMC-derived weights for the class corresponding to $t$. 

Define $\alpha(\cdot)$ as:
\begin{equation}
    \alpha(t, k) = 
    \begin{cases}
        \beta & \text{if $k = \lfloor t \cdot |K| \rfloor$} \\
        1 & \text{else}
    \end{cases}
\label{eq:alpha_t}
\end{equation}

\noindent for $\beta \geq 1$. Here, each value of $t$ corresponds to a context where the associated class is prioritized which is reflected in the loss of Eq.~\ref{eq:alpha_t}. We set the endpoint models for the bridge mode to correspond to the same initial mode weights but allow the weights at each endpoint to float during the BMC optimization. This effectively results in two different endpoints by the end of training. We follow the same training process as the original BMC with respect to data sampling. The results provided in Fig.~\ref{fig:exp1_exp2} show that the BMC successfully shifts the performance profile by improving the accuracy for the up-weighted class relative to an unweighted BMC baseline.

\subsection{Long-tail Robustness}
To enable the BMC to model context for long-tail robustness, we update the data that is used for training the BMC after it is sampled.  A corruption, $\psi$, is applied to each mini-batch $B_i$. Every element $x_j \in B_i$ is updated with $\psi(x_j, t)=x'_j$ before it is used for training and where $x \sim \mathcal{D}$.  We normalize the corruption severity to range between $0$ and $1$ such that it maps linearly to $t$. 
Our experiment adds Gaussian noise with $\mu=0$ and $\sigma = \gamma t$ (for scaling factor $\gamma$ which bounds the perturbation magnitude). The endpoints are allowed to float and other details of the training follow the standard BMC procedure described above. The results are shown in Fig.~\ref{fig:exp1_exp2}, and demonstrate the BMC not only smoothly interpolates between the range of uncorrupted and corrupted contexts, but achieves small gains in a subset of contexts. Since the real-world allows for many simultaneous corruptions and robustness challenges, we extend our approach to the two-dimensional case where we can generalize the BMC to account for combinations corruptions during training/evaluation. Details of this approach and results are found in Appendix~\ref{app:planar}.

\subsection{Contextual Distributional Shift}
\begin{figure}[t]
\centering
  \includegraphics[width=.47\linewidth]{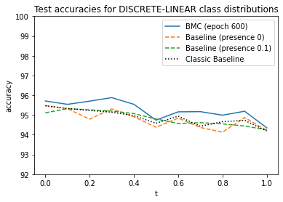}
  \captionof{figure}{\textbf{Contextual Distribution Shift: } We vary $P(Y=c|t)$ linearly for each class along $t$ with different probability distributions above and below $t=0.5$.  The BMC successfully accounts for this variation and achieves small gains over the baseline one-size-fits-all model.}
  \label{fig:exp3}
\end{figure}

Lastly, we experiment with contextual distributional shift by altering the class distribution of the data based on sampled values of $t$. Here, minibatch $B_i$ is sampled according to a defined distribution of target classes based on $t$ (i.e., $P(Y|t)$). We experiment specifically with the case where all classes are sampled with $P(Y=c|t) = (1-2t)(c - 0.5 \cdot N)/N + 0.5$, where $N$ is the number of classes and all classes are equally likely at $t=0.5$.  The results are shown in Fig.~\ref{fig:exp3} and illustrate that the BMC is able to compensate well for the changes in class distribution as $t$ varies and even achieve a small gain over baseline models in this case.

\section{Discussion}
Given the need for ML models to work well in a range of contexts, we consider here a shift away from one-size-fits-all approaches toward context-adaptive methods.  We introduced three interpretations of context and made novel extensions to Bridge-Mode Connectivity to train context-sensitive models.  Our results show that we can train Bridge-Mode curves with these various context perspectives and show that performance across contexts is maintained in all cases.  We hypothesize that the marginal increase in performance is due to the lack of diversity in the weights along the curve (even after the context-sensitive training). This lack of diversity was also noted in the original formulation by~\cite{garipov2018loss} and remains an open topic of exploration in the context of training Bridge-Mode ensembles. 

The models sampled from the context-adaptive BMC provide an added degree of trust relative to one-size-fits-all methods.  This is taken from the fact that the construction of the BMC loss and/or optimization ensures that weights sampled from the BMC for a specific context have been directly optimized for that context in terms of the perceived risks or properties of the data.  This cannot be said for single, static models which are optimized over all (or none) of the contexts without any guarantee of performance for any specific point on that continuum. This may be an important consideration when attempting to certify the safety of ML models for deployment.

Overall, we believe this work demonstrates novel methods for utilizing the BMC towards achieving context-adaptive models and lays the foundation for future work to focus on improving the specialization of models to their associated context along the BMC.

{\small
\bibliographystyle{ieee_fullname}
\bibliography{egbib}
}

\clearpage
\appendix
\section{Model architecture and training details}

For our main experiments, we utilize the ResNet18 architecture~\cite{he2016deep} as the seed model for the BMC training.  We train all endpoint models using standard stochastic gradient descent. The learning rate is initialized at 0.1, and is then adjusted at each epoch based on cosine annealing. Further, the momentum is set to 0.9, the weight decay is set to 0.0005, and we use Nesterov momentum. We train for 200 epochs with a batch size of 128. Images are also normalized prior to corruption, and during training, random flipping and cropping are applied.

BMC models are also trained with stochastic gradient descent, but for 600 epochs. We use a Bezier curve with one bend for our curve parameterization, and weight decay of 0.0005, and an initial learning rate of 0.015. The learning rate varies according to Equation~\ref{eq:bmc_lr}:

\begin{align}
    \text{Learning Rate} = \begin{cases}
    r & \text{if } \alpha \leq 0.5 \\
    (1 - ((\alpha - 0.5) \cdot 2.5 \cdot 0.99))r & \text{if } 0.5 \leq \alpha \leq 0.9 \\
    0.01r & \text{otherwise}
    \end{cases}
    \label{eq:bmc_lr}
\end{align}
where $r$ is the initial learning rate; i.e. $r=0.015$ in our experiments.

\section{Planar Model - 2D BMC}
\label{app:planar}
Extending from the initial formulation in~\cite{landscape2019}, we train a planar model composed of the following trainable elements:
\begin{itemize}
    \item ${w}_0 \in \mathbb{R}^m$: the vectorized weights of a pre-trained base model, where $m \in \mathbb{N}$ is the number of parameters for the base model, 
    \item ${w}_1 \in \mathbb{R}^m$: a vector of weights corresponding to the first corruption,
    \item ${w}_2 \in \mathbb{R}^m$: a vector of weights corresponding to the second corruption,
    \item $s \in \mathbb{R}$: a scaling factor for ${w}_1$ and ${w}_2$
\end{itemize}

\begin{figure}
    \centering
    \includegraphics[width=\linewidth, keepaspectratio]{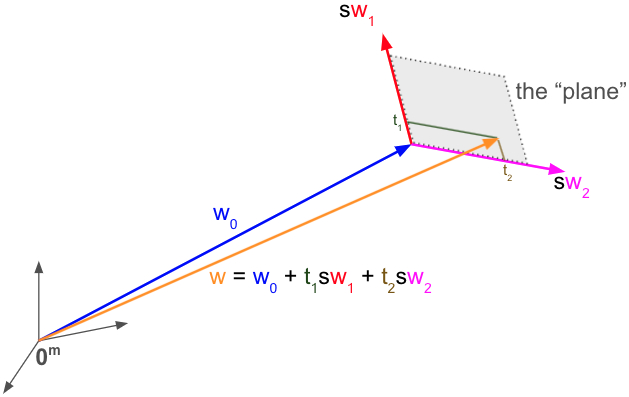}
    \caption{The planar model components $w_0, w_1, w_2, s$ visualized in weight-space $\mathbb{R}^m$, where $m$ is the number of parameters in the base model architecture. A corruption point $(t_1, t_2) \in [0, 1]^2$ determines (1) the corruption levels of the input images (where $t_c = 0$ is no corruption and $t_c = 1$ is severe corruption for corruptions $c = 1, 2$) and (2) the model weights $w$ used to classify the corrupted images.}
    \label{fig:planar_model}
\end{figure}

We parameterize corruptions by $t_c \in [0, 1]$ for corruptions $c = 1, 2$, where $t_c = 0$ corresponds to no corruption and $t_c = 1$ corresponds to severe corruption. During training and testing, we selected a corruption point $(t_1, t_2) \in [0, 1]^2$, corrupted the input images accordingly, and classified the corrupted images with a model instantiated with weights ${w}$, where
$$
{w} = {w}_0 + t_1s{w}_1 + t_2s{w}_2.
$$
Conceptually, the set of weights $\{{w} | (t_1, t_2) \in [0, 1]^2\}$ provided by the planar model corresponds a parallelogram on a 2D plane in $\mathbb{R}^m$ (Figure~\ref{fig:planar_model}).

During training, we accumulated loss across 50 randomized $(t_1, t_2) \in [0, 1]^2$ points before performing stochastic gradient descent. After each epoch, we evaluated the accuracy of the planar model at all points in the set $\{(t_1, t_2) | t_1, t_2 \in \{0, 0.1, 0.2, ..., 1\} \}$ and terminated training when the average accuracy did not increase for 20 epochs.

During testing, we evaluated the accuracy of the planar model at all points in the set $\{(t_1, t_2) | t_1, t_2 \in \{0, 0.1, 0.2, ..., 1\} \}$.

\begin{figure}
    \centering
    \includegraphics[width=\linewidth, keepaspectratio]{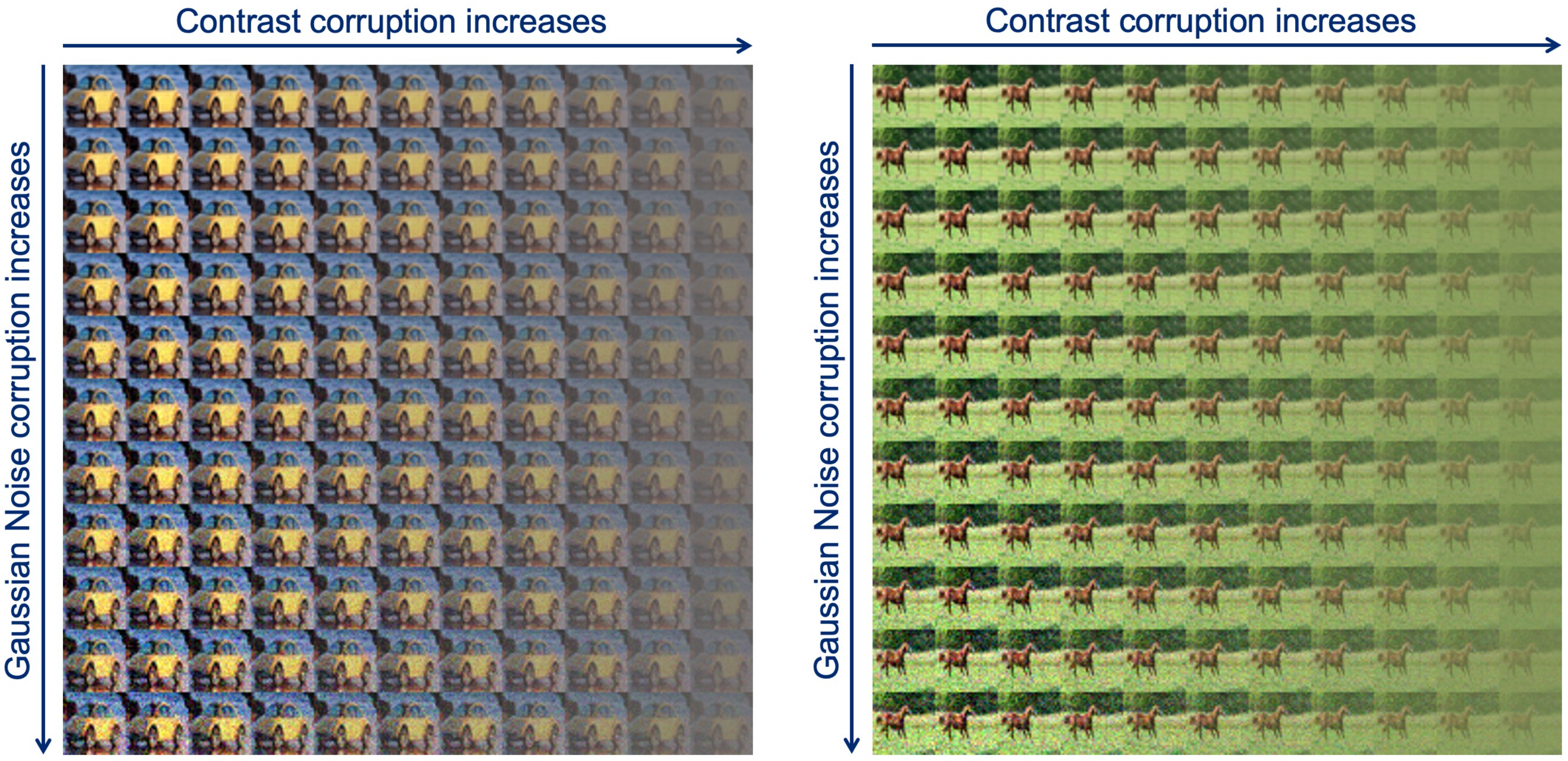}
    \caption{Examples of images that are corrupted by Gaussian noise and contrast corruptions. The upper-left image has no corruptions, while Gaussian noise increases from top to down and contrast corruption increases from left to right. The lower-right corner is thus severely corrupted by both corruptions. This layout corresponds to the accuracy values presented in subsequent figures.}
    \label{fig:sample_corruptions}
\end{figure}

We constructed and evaluated the planar model using Gaussian noise and contrast image corruptions, with two examples shown in Figure~\ref{fig:sample_corruptions}. For our base model, we used a network architecture which contains 4 convolutional layers (with [32, 32, 64, 64] channels)and  3 fully connected layers (with [512, 512, 10] nodes) with ReLU activations for a total of 2,466,858 trainable parameters (i.e. $m = 2466858$).

\begin{figure}
    \centering
    \includegraphics[width=\linewidth, keepaspectratio]{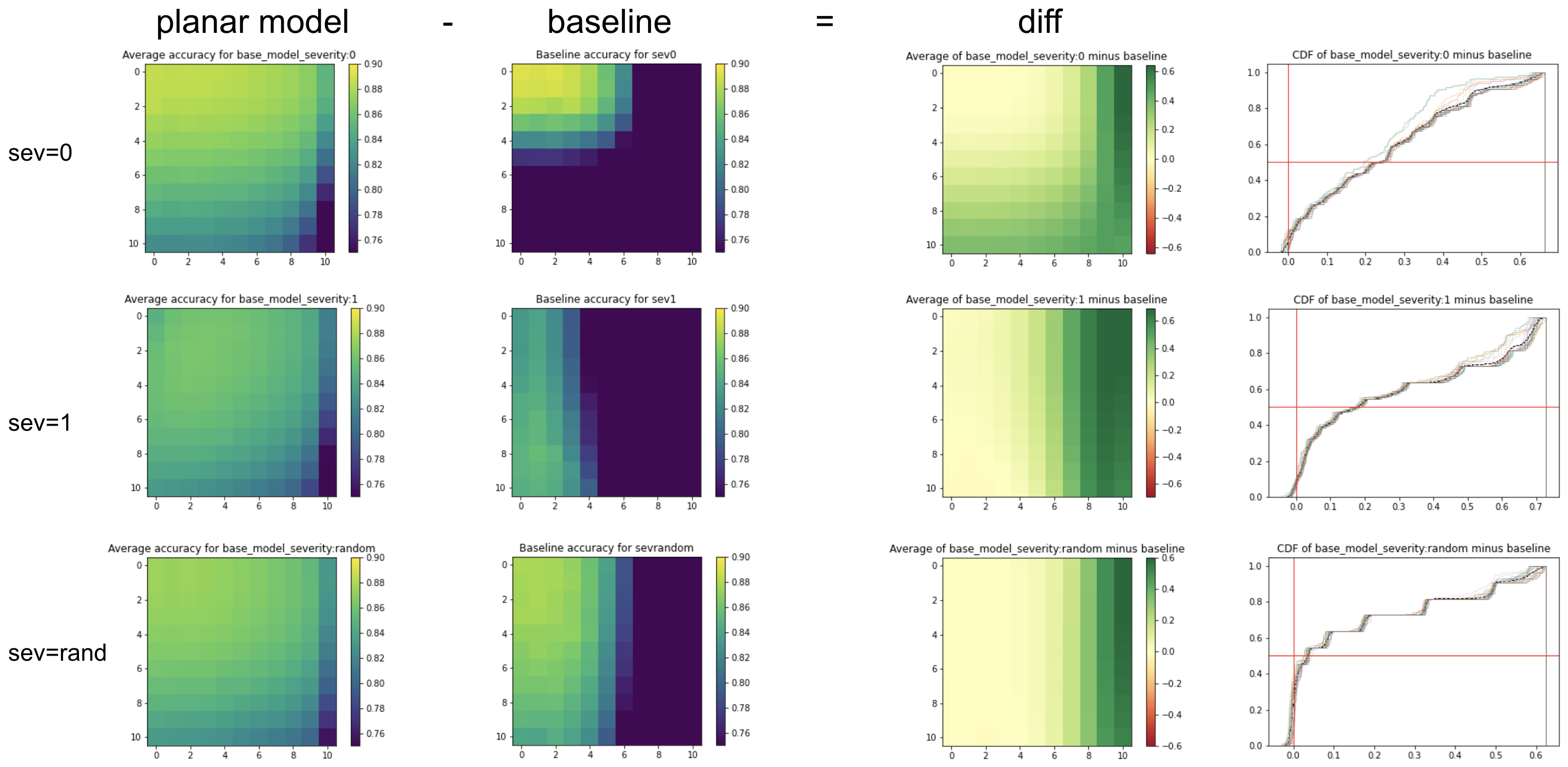}
    \caption{Accuracy of the planar models and the base models on corrupted images, where in each grid, the upper-left is evaluated on images without corruptions and lower-right is evaluated on highly corrupted images. For base models, the one trained only on clean images has a quick drop-off in accuracy as images are corrupted, while base models trained with Gaussian noise corruptions do well on all severities of Gaussian noise corruption, with the planar model mostly improving accuracy on contrast-corrupted images.}
    \label{fig:baselines}
\end{figure}

Figure~\ref{fig:baselines} illustrates that in the 2D case, we're still able to train the Bridge-Mode to achieve more uniform performance across all corruptions compared to the baseline.  In the case where both the baseline and BMC were trained on corrupted data (bottom row of Fig.~\ref{fig:baselines}), the BMC improves measurably over the baseline indicating better context-specific adaptation for the more severe noise and contrast corruptions.

\clearpage

\end{document}